\documentclass{article}

     \usepackage[preprint, nonatbib]{neurips_2019}

\usepackage[utf8]{inputenc} 
\usepackage[T1]{fontenc}    
\usepackage{hyperref}       
\usepackage{url}            
\usepackage{booktabs}       
\usepackage{amsfonts}       
\usepackage{nicefrac}       
\usepackage{microtype}      

\usepackage{amsfonts, amsmath, amssymb, color, graphicx}

\usepackage{url}
\newtheorem{lem}{Lemma}

\newtheorem{thm}{Theorem}

\usepackage{tikz}

 \usetikzlibrary{shapes, arrows, positioning, decorations.markings}

\definecolor {processblue}{cmyk}{0.96,0,0,0}
\tikzstyle{int}=[draw, fill=blue!20, minimum size=2em]
\tikzstyle{init} = [pin edge={to-,thin,black}]

\usetikzlibrary{shapes}
\usetikzlibrary{fit}
\usetikzlibrary{chains}
\usetikzlibrary{arrows}

\tikzstyle{plate} = [draw, rectangle, rounded corners, fit=#1]
\tikzstyle{wrap} = [inner sep=0pt, fit=#1]

\title{MixUp as Directional Adversarial Training}

\author{%
  Guillaume P. Archambault\\
School  of Electrical Engineering and Computer Science, 
  University of Ottawa\\
  Ottawa, ON K1N 6N5, 
 Canada\\ 
  \texttt{gperr050@uottawa.ca} \\
  \And
  Yongyi Mao\\
  School  of Electrical Engineering and Computer Science, 
  University of Ottawa\\
  Ottawa, ON K1N 6N5,  
 Canada\\ 
  \texttt{yymao@eecs.uottawa.ca} \\
  \AND
  Hongyu Guo\\
   National Research Council of Canada\\
1200 Montreal Road, Ottawa, ON K1A 0R6, Canada\\
  \texttt{hongyu.guo@nrc-cnrc.gc.ca} \\
   \And
  Richong Zhang \\
School of Computer Science, Beihang University\\
37 Xueyuan Rd., Haidian Dist., Beijing 100191, China\\
\texttt{zhangrc@act.buaa.edu.cn} \\
}

\begin{document}

\maketitle

\begin{abstract}
In this work, we explain the working mechanism of MixUp in terms of adversarial training. We introduce a new class of adversarial training schemes, which we refer to as directional adversarial training, or DAT. In a nutshell, a DAT scheme perturbs a training example in the direction of another example but keeps its original label as the training target. We prove that MixUp is equivalent to a special subclass of DAT, in that it has the same expected loss function and corresponds to the same optimization problem asymptotically. This understanding not only serves to explain the effectiveness of MixUp, but also reveals a more general family of MixUp schemes, which we call Untied MixUp. We prove that the family of Untied MixUp schemes is equivalent to the entire class of DAT schemes. We establish empirically the existence of Untied Mixup schemes which improve upon MixUp.

\end{abstract}

\section{Introduction}

The success of neural network models in the modern paradigm of deep learning often requires the construction of complicated networks with a large number of parameters (see, e.g., \cite{resNet, wideResNet, Bert}). Such network models thus often have overwhelmingly high capacities.  Although it is still unclear to date what makes a neural network generalize well\cite{RethinkingGeneralization, CloserLookAtMemorization_Bengio}, the high capacities of these models are observably prone to overfitting and effective regularization techniques are highly demanded in the training of these models. 

Beyond the classical regularization techniques such as weight decay \cite{WeightDecay} or dropout \cite{dropout}, recent research has been paving the ways in two new directions.  

One direction is adversarial training\cite{szegedy2013intriguing}, in which a training example is perturbed under a certain designed strategy in the data space and the perturbed example is trained using its original label. This allows the model to consider some unseen region near the data point as having the same label, thereby further constraining the model and pushing it towards better generalization.  Such a technique has shown to be effective and has attracted active research interest (see, e.g.,  \cite{goodFellowAdv,miyato2016adversarial, athalye2018obfuscated, shaham2018understanding, he2018decision}).

Another direction is known as ``MixUp''\cite{zhang2017mixup}, in which one synthesizes a new training example by interpolating a pair of training examples and using a weighted combination of their respective labels as the training objective.  Despite its appealing effectiveness demonstrated in recent literature \cite{zhang2017mixup, adaMixUp, manifoldMixUp}, the working mechanism of MixUp has not been well understood to date. The authors of \cite{adaMixUp} suggest viewing MixUp as imposing certain ``local linearity'' on the model using points outside of the data manifold. Though this is correct, it still does not fully explain why MixUp works. 

This research is motivated by a curiosity to better understand the working of MixUp. In this work, we discover that the working principle of MixUp is in fact very similar to that of adversarial training. More precisely, we show that MixUp can be seen as ``equivalent'', in a particular sense, to a new family of adversarial training schemes, which we call {\em Directional Adversarial Training}, or DAT. In DAT, the strategy of perturbing  examples does not follow the conventional approaches, e.g, that of \cite{goodFellowAdv}.  Instead, to perturb an example $x$, DAT picks a random example $x'$,  draws a random fractional number $\lambda$ from a prescribed distribution, and perturbs $x$ towards $x'$ by $(1-\lambda)$ fraction of the distance between $x$ and $x'$. 

The consequence of establishing the equivalence between MixUp and DAT is two-fold. First it allows an understanding of MixUp from the viewpoint of adversarial training. On one hand, this viewpoint at least partially explains the effectiveness of MixUp. On the other hand, it also allows the insights developed in literature of adversarial training to assist further developing MixUp, and vice versa. The second consequence of this equivalence is that it shows that MixUp is only equivalent to a subclass of  DAT. Two questions then naturally arise. 
\begin{enumerate}
\item
What are the other members of the DAT family that do not correspond to MixUp? 
\item Can these members be employed in regularization schemes as effective as, or even better than, MixUp?
\end{enumerate}

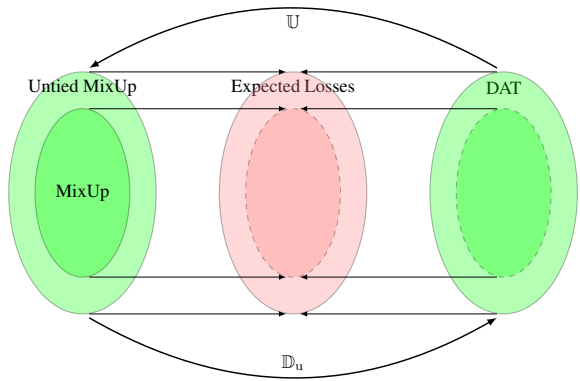
\begin{figure}
\begin{center}
\scalebox{0.7}{
\begin{tikzpicture}[-latex ,auto ,node distance =2 cm and 2 cm ,on grid ,
semithick ,
state/.style ={ circle ,top color =white , bottom color = processblue!20 ,
draw, processblue , text=black , minimum width =0.25cm, text width=0.3cm,},
box/.style ={rectangle ,top color =white , bottom color = processblue!20 ,
draw, processblue , text=blue , minimum width =0.5cm , minimum height = 0.5cm},
highbox/.style ={rectangle ,top color =white , bottom color = processblue!20 ,
draw, processblue , text=blue , minimum width =1.7cm , minimum height = 1.6cm},
neuron/.style ={rectangle ,top color =white , bottom color = red!20 ,
draw, red , text=red , minimum width =0.5cm , minimum height = 0.5cm, rounded corners},
triangle/.style = {top color =white , bottom color = processblue!20 ,
draw, processblue , text=blue, regular polygon, regular polygon sides=3, minimum size=0.5cm, draw },
node rotated/.style = {rotate=270},
    border rotated/.style = {shape border rotate=270}]
\definecolor{cof}{RGB}{219,144,71}
\definecolor{pur}{RGB}{186,146,162}
\definecolor{greeo}{RGB}{91,173,69}
\definecolor{greet}{RGB}{52,111,72}


\node[ellipse ,
draw, black, fill=green, text=red , minimum width =2.8cm , minimum height = 4.6cm, rounded corners,opacity=0.3] (umix) at (0,0) {};

\node[ellipse ,
draw, black, fill=green, text=black , minimum width =1.8cm , minimum height = 3.2cm, rounded corners,opacity=0.3] at (0,0) {};

\node[] at (0, 0){MixUp};
\node[] at (0, 2){Untied MixUp};
\node[] at (8, 2){DAT};
\node[] at (4, 2){Expected Losses};


\node[ellipse ,
draw, black, fill=red!50, text=red , minimum width =2.8cm , minimum height = 4.6cm, rounded corners,opacity=0.3] (expect)at (4,0) {};

\node[ellipse ,
draw, black, dashed, fill=red!50, text=red , minimum width =1.8cm , minimum height = 3.2cm, rounded corners,opacity=0.3] at (4,0) {};


\node[ellipse ,
draw, black, fill=green, text=red , minimum width =2.8cm , minimum height = 4.6cm, rounded corners,opacity=0.3] (dat) at (8,0) {};

\node[ellipse ,
draw, black, dashed, fill=green, text=red , minimum width =1.8cm , minimum height = 3.2cm, rounded corners,opacity=0.3] at (8,0) {};

\node[] (mixTop) at (0, 1.6){};
\node[] (mixBot) at (0, -1.6){};
\node[] (expectSmallTop) at (4, 1.6){};
\node[] (expectSmallBot) at (4, -1.6){};
\node[] (datSmallTop) at (8, 1.6){};
\node[] (datSmallBot) at (8, -1.6){};

\node[] (umixTop) at (0, 2.3){};
\node[] (umixBot) at (0, -2.3){};
\node[] (expectBigTop) at (4, 2.3){};
\node[] (expectBigBot) at (4, -2.3){};
\node[] (datTop) at (8, 2.3){};
\node[] (datBot) at (8, -2.3){};

\draw[thin] (mixTop) -- (expectSmallTop);
\draw [thin] (mixBot) -- (expectSmallBot);
\draw[thin] (datSmallTop) -- (expectSmallTop);
\draw [thin] (datSmallBot) -- (expectSmallBot);

\draw[thin] (umixTop) -- (expectBigTop);
\draw [thin] (umixBot) -- (expectBigBot);
\draw[thin] (datTop) -- (expectBigTop);
\draw [thin] (datBot) -- (expectBigBot);


\node [](umix_low) at (0, -2.3){};
\node [](dat_low) at (8, -2.3){};
\node [](umix_high) at (0, 2.3){};
\node [](dat_high) at (8, 2.3){};

\path [thick](umix_low) edge [out=-30, in=210] node[]{${\mathbb D}_{\rm u}$}(dat_low);
\path [thick](dat_high) edge [out=150, in=30] node[]{${\mathbb U}$}(umix_high);

%
%

\end{tikzpicture}}
\caption{\label{fig:diagram} The relationship between MixUp, DAT and Untied MixUp.}
\end{center}
\end{figure}

Question 1 is fully resolved in this work. We show that there is a more general family of MixUp schemes, which we call Untied MixUp and that every member of DAT is equivalent to some member(s) of the Untied MixUp family.  
The relationship between MixUp, DAT, and Untied MixUp is shown in Figure \ref{fig:diagram}. 

We also have  investigated Question 2 experimentally. We perform an ad hoc search in the space of Untied MixUp schemes and experimentally evaluate their performances. Our results suggest that, indeed, some Untied MixUp schemes can be confidently claimed as more effective than the best known MixUp.  

Finally we note that this paper has another contribution. Conventionally MixUp is only applicable to baseline models defined using the cross entropy loss. All analytical results we develop in this paper are in fact applicable to a much wider model family, beyond those defined using cross-entropy loss. In a sense, we have presented a generalization of MixUp in another dimension. 

Necessary proofs of our results are included in Supplementary Materials.


\section{MixUp as Directional Adversarial Training}

\subsection{Classification Models with Target-Linear Loss Functions}

Consider a standard classification problem, in which one wish to learn a classifier that predicts the {\em class label} for an {\em object}.

Formally, let ${\cal X}$ be a vector space in which the objects of interest live and let ${\cal Y}$ be the set of all possible labels of these objects.
The set of training examples will be denoted by ${\cal D}$, identified with a subset of ${\cal X}$. We will use $t(x)$ to denote the true label of $x$.  
Let $F$ be a neural network function, parameterized by $\theta$, which  maps ${\cal X}$ to another vector space ${\cal Z}$. 
Let $\varphi:{\cal Y} \rightarrow {\cal Z}$ be a function that maps a label in ${\cal Y}$ to an element in ${\cal Z}$ such that for any $y, y' \in {\cal Y}$,  if $y\neq y'$, then $\varphi(y) \neq \varphi (y')$. Usually one would also require $\varphi(y)$ and $\varphi(y')$ to be sufficiently apart under some metric in ${\cal Z}$. But this is not a primary concern of this paper. 

In the space ${\cal Z}$, we refer to $F(x)$  as the {\em model's prediction} for $x$ and $\varphi(t(x))$ as the {\em training target} of $x$. 

Let $\ell: {\cal Z} \times {\cal Z}\rightarrow {\mathbb R}$ be a {\em loss function}, using which one defines an {\em overall loss function} as 
\begin{equation}
\label{eq:totalLoss}
{\cal L}: = 
\frac{1}{|{\cal D}|}
\sum\limits_{x\in {\cal D}} \ell \left(F(x), \varphi(t(x))\right)
\end{equation}
Here we have taken the notational convention that the second argument of $\ell$ represents the target whereas the first represents the model's prediction.  

In this setting, the learning problem is then formulated as minimizing ${\cal L}$ with respect to its parameter $\theta$. 

We now single out a family of classification models.  To that end, we say that the loss function $\ell (z', z)$ is {\em target-linear} if for any scalars $\alpha$ and $\beta$,
$\ell(z', \alpha z_1+\beta z_2) = \alpha \ell(z', z_1) + \beta \ell (z', z_2).$

As examples, we next give two families of models that have target-linear loss functions. 

\noindent{\bf Models with cross-entropy loss.} Let ${\cal Z}$ be ${\cal Z}_{\rm p}$, namely, the family of all distributions over ${\cal Y}$. Each $a\in {\cal Y}$ can be associated with a trivial  distribution $\delta_a \in {\cal Z}_{\rm p}$ defined by $\delta_a(y)=1$ if and only if $y=a$. Viewed as a vector, $\delta_a$ is simply the ``one-hot vector'' with $1$ on the $a^{\rm th}$ location.  The association of $a\in {\cal Y}$ with $\delta_a\in {\cal Z}_{\rm p}$ essentially defines the function $\varphi$. That is, $\varphi (a) = \delta_a$. In this setting, the loss function $\ell$ can be taken as the  cross-entropy loss $\ell_{\rm CE}$, defined by
\[
\ell_{\rm CE}(q,  p):  = \sum\limits_{y\in {\cal Y}} p(y)\log q(y).
\]
for any two $p, q \in {\cal Z}_{\rm p}$. 


\noindent {\bf Models with negative-cosine loss.} Let ${\cal Z}:={\cal Z}_{\rm u}$, namely the set of all unit length vectors in some vector space ${\mathbb R}^M$. Each $y\in {\cal Y}$ is mapped to a distinct vector in ${\cal Z}_{\rm u}$ under $\varphi$.  
In this setting, let the loss function $\ell$ be $\ell_{\rm NC}$, defined as
\[
\ell_{\rm NC}(z' , z):= -z'^{\rm T} z 
\]
Note that $\ell_{\rm NC}(z' , z)$ is essentially the negative cosine similarity between $z$ and $z'$, which we call the ``negative-cosine loss''. 

\begin{lem} The $\ell_{\rm CE}$ and  $\ell_{\rm NC}$ are both target-linear.
\end{lem}

In the paper, we restrict our discussion to the models in which the loss function $\ell$ is target-linear\footnote{In fact, all theoretical results developed in this paper can be extended to models with ``target-affine'' loss functions, where the ``target-affine'' property extends the definition of target-linearity in a natural way.}.

Much of the development in this paper concerns drawing objective pairs $(x, x')$ from ${\cal D}\times {\cal D}$. For later use, a distribution $Q$ on ${\cal D}\times{\cal D}$ will be called {\em exchangeable}, or {\em symmetric}, if for any $(x, x')\in {\cal D}\times{\cal D}$, $Q(x, x')=Q(x', x)$.  Throughout the paper, we will assume that there is a symmetric distribution $Q$ on ${\cal D}\times{\cal D}$. In practice $Q$ is often taken as the uniform distribution. 

Throughout the paper, we will use capitalized letters, e.g., $X$, to denote random variables, and their lower cased counterparts, e.g., $x$, to denote values that the random variables make take. Any sequence,  $(a_1, a_2, \ldots, a_n)$ will be denoted by $a_1^n$. Likewise $(A_1, A_2, \ldots, A_n)$ will be denoted by $A_1^n$, and a sequence of object pairs $\left((x_1, x'_1), (x_2, x'_2), \ldots, (x_K, x'_K)\right)$ denoted by $(x, x')_1^K$. 

Suppose that $(x, x')_1^K$ is a length-$K$ sequence of object pairs drawn from ${\cal D}\times {\cal D}$; the sequence will be called symmetric if the empirical distribution of $(X, X')$ observed in $(x, x')_1^K$ is symmetric. Alternatively put, the sequence $(x, x')_1^K$ is said to be symmetric if for every $(a, b)\in {\cal D}\times {\cal D}$, the number of occurrences of $(a, b)$ in the sequence is equal to that of $(b, a)$.

For any value $a\in [0, 1]$, we will use $\overline{a}$ as a short notation for $1-a$.


\subsection{MixUp}

For any $x, x' \in {\cal D}$ and any $\lambda \in [0, 1]$, denote 

\begin{equation}
\ell^{\rm Mix}(x, x', \lambda): = 
\ell 
\left(
F(\lambda x + \overline{\lambda}x'), \lambda\varphi(t(x)) + \overline{\lambda}\varphi(t(x'))
\right)
\end{equation}

Let $P^{\rm Mix}$ be a distribution over $[0, 1]$, and $K$ be a positive integer.  In MixUp,  a sequence $(x, x')_1^K: =\left((x_1, x'_1), (x_2, x'_2), \ldots, (x_K, x'_K)\right)$ of example pairs are drawn i.i.d. from $Q$, and a sequence $\lambda_1^K:=\left(\lambda_1, \lambda_2, \ldots, \lambda_K\right)$ of values are drawn i.i.d. from $P^{\rm Mix}$.  Define
\begin{eqnarray}
\label{eq:Lmix}
{\cal L}^{\rm Mix}\left((x, x' )_1^K, \lambda_1^K \right) & :=  &
\frac{1}{K} \sum\limits_{k=1}^K 
\ell^{\rm Mix}(x_k, x'_k, \lambda_k) \\
\label{eq:expLmix}
{\cal L}^{\rm Mix}\left((x, x' )_1^K\right) & := & 
{\mathbb E}_{\lambda_1^K \stackrel{\rm iid}{\sim} P^{\rm Mix}}
{\cal L}^{\rm Mix}\left((x, x' )_1^K, \lambda_1^K \right)
\end{eqnarray}
Above we have overloaded the notation ${\cal L}^{\rm Mix}$. One should distinguish its meaning according to the argument it takes.  In MixUp, the overall loss function is taken as ${\cal L}^{\rm Mix}\left((x, x' )_1^K, \lambda_1^K \right)$ for a random choice of $((x, x')_1^K, \lambda_1^K)$ as defined above and
is minimized with respect to network parameter $\theta$.  In MixUp, we refer to $P^{\rm Mix}$ as the {\em mixing policy}.


\begin{lem}
\label{lem:mixUpConverge1}
For any fixed infinite sequence $(x, x')_1^\infty$ and any infinite sequence of i.i.d. random variables $\Lambda_1^\infty$ drawn from $P^{\rm Mix}$, let 
${\cal L}^{\rm Mix}\left((x, x' )_1^K, \Lambda_1^K \right)$ be defined according to (\ref{eq:Lmix}), with the first $K$ elements of $(x, x')_1^\infty$ and the first $K$ elements of $\Lambda_1^\infty$ as input. Let
\[
\delta_{\rm Mix}: = \max_{(x, x') \in {\cal D} \times {\cal D}}  \sup_{(\lambda, \lambda') \in [0, 1]\times [0, 1]} \left| \ell^{\rm Mix}(x, x', \lambda) - \ell^{\rm Mix}(x, x', \lambda') \right| .
\]
Then  for any $\epsilon >0$, 
$
{\rm Pr}\left[
\left|
{\cal L}^{\rm Mix}\left(
(x, x')_1^K, \Lambda_1^K
\right) -  {\cal L}^{\rm Mix}\left(
\left(x, x'\right)_1^K
\right) 
\right| \ge \epsilon
\right]
<
2 \exp \left( -  \frac{2\epsilon^2}{\delta_{\rm Mix}^2} \cdot K\right).
$
\end{lem}

%
%

By this lemma, as $K$ increases, 
${\cal L}^{\rm Mix}\left(
(x, x')_1^K, \Lambda_1^K
\right)$ 
converges to   
${\cal L}^{\rm Mix}\left(
\left(x, x'\right)_1^K
\right) 
$ in probability. 


\subsection{Directional Adversarial Training (DAT)}

For any $x\in {\cal D}$, $x' \in {\cal X}$ and $\lambda\in [0, 1]$, we denote
\begin{equation}
\ell^{\rm DAT}(x \rightarrow x', s): = \ell \left(F(sx+\overline{s}x'), \varphi(t(x))\right)
\end{equation}

Let $P^{\rm DAT}$ be a distribution over $[0, 1]$ and let $K$ be a positive integer.  In Directional Adversarial Training, or DAT,  a sequence $(x, x')_1^K: =\left((x_1, x'_1), (x_2, x'_2), \ldots, (x_K, x'_K)\right)$ of example pairs are drawn i.i.d. from $Q$, and a sequence $\lambda_1^K:=\left(\lambda_1, \lambda_2, \ldots, \lambda_K\right)$ of values are drawn i.i.d. from $P^{\rm DAT}$.  Define
\begin{eqnarray}
\label{eq:Ldat}
{\cal L}^{\rm DAT}\left((x, x' )_1^K, \lambda_1^K\} \right) & : = &
\frac{1}{K}\sum\limits_{k=1}^K 
\ell^{\rm DAT} 
\left(x_k \rightarrow x'_k, \lambda_k\right)\\
\label{eq:expLdat}
{\cal L}^{\rm DAT}\left((x, x' )_1^K\right) &:= &
{\mathbb E}_{\lambda_1^K\stackrel{\rm iid}{\sim}P^{\rm DAT}} {\cal L}^{\rm DAT}\left((x, x' )_1^K, \lambda_1^K\right)
\end{eqnarray}

In DAT, the overall loss function is taken as ${\cal L}^{\rm DAT}\left((x, x' )_1^K, \lambda_1^K \right)$ for a random choice of $((x, x')_1^K, \lambda_1^K)$ as defined above and
is minimized with respect to network parameter $\theta$. Note that this loss function indeed defines an adversarial training scheme, since the training example $x$ is moved towards the direction of $x'$ but its label  is kept.  

In DAT, we refer to $P^{\rm DAT}$ as the {\em adversarial policy}.


\begin{lem}
\label{lem:DATConverge1}
For any fixed infinite sequence $(x, x')_1^\infty$ and any infinite sequence of i.i.d. random variables $\Lambda_1^\infty$ drawn from $P^{\rm DAT}$, let 
${\cal L}^{\rm Mix}\left((x, x' )_1^K, \Lambda_1^K \right)$ be defined according to (\ref{eq:Ldat}), with the first $K$ elements of $(x, x')_1^\infty$ and the first $K$ elements of $\Lambda_1^\infty$ as input. Let
\[
\delta_{\rm DAT}: = \max_{(x, x') \in {\cal D} \times {\cal D}}  \sup_{(\lambda, \lambda') \in [0, 1]\times [0, 1]} \left| \ell^{\rm DAT}(x, x', \lambda) - \ell^{\rm DAT}(x, x', \lambda') \right| .
\]
Then  for any $\epsilon >0$, 
$
{\rm Pr}\left[
\left|
{\cal L}^{\rm Mix}\left(
(x, x')_1^K, \Lambda_1^K
\right) -  {\cal L}^{\rm Mix}\left(
\left(x, x'\right)_1^K
\right) 
\right| \ge \epsilon
\right]
<
2 \exp \left( -  \frac{2\epsilon^2}{\delta_{\rm DAT}^2} \cdot K\right).
$
\end{lem}

The proof of this lemma follows in exactly the same way as that of Lemma \ref{lem:mixUpConverge1}. It follows from this lemma that as $K$ increases, 
${\cal L}^{\rm DAT}\left(
(x, x')_1^K, \Lambda_1^K
\right)$ 
converges to   
${\cal L}^{\rm DAT}\left(
\left(x, x'\right)_1^K
\right) 
$ in probability. 


\subsection{Relationship between MixUp and DAT}

We first inspect the loss function $\ell^{\rm Mix}$ and $\ell^{\rm DAT}$ and the following lemma follows immediately from the target-linearity of the underlying loss function $\ell$. 

\begin{lem}
\label{lem:key}
For any $(x, x') \in {\cal D} \times {\cal D}$ and any $\lambda \in [0, 1]$,
\begin{eqnarray*}
\ell^{\rm Mix}(x, x', \lambda)  & = & \lambda \ell^{\rm DAT} (x \rightarrow x', \lambda) + \overline{\lambda} \ell^{\rm DAT} (x' \rightarrow x,\overline{\lambda}) \\
& =  &
{\mathbb E}_{W|\lambda}  \left\{
W \cdot  \ell^{\rm DAT} (x \rightarrow x', \lambda) + (1-W) \cdot \ell^{\rm DAT} (x' \rightarrow x,\overline{\lambda})
\right\}
\end{eqnarray*}
where $W|\lambda$ is a ${\rm Bernoulli}(\lambda)$  (also written as ${\rm Ber}(\lambda)$) random variable, namely a $\{0, 1\}$-valued random variable that takes value $1$ with probability $\lambda$. 
\end{lem}

That is, a single training case with deterministic mixing policy $\lambda$ in MixUp has the {\em average effect} of two training cases in DAT, where the averaging is over random draws of the two training cases governed by a ${\rm Ber}(\lambda)$ random variable. This simple lemma thus provides a fundamental connection between MixUp and DAT.

To go beyond MixUp with deterministic policies, let ${\cal P}$ denote the space of all distributions on $[0, 1]$. Thus, ${\cal P}$ is the space of all mixing policies for MixUp as well as the space of all adversarial policies for DAT. Let ${\mathbb D}$ be a mapping from ${\cal P}$ to ${\cal P}$ defined as follows. For any distribution $p\in {\cal P}$, ${\mathbb D}(p)$ is the distribution $p'$ defined by
\[
p'(\lambda):= \lambda \left(p(\lambda) + p(1-\lambda) \right)
\] 
for every $\lambda \in [0, 1]$.  We note that it can be easily verified that $p'$ defined this way satisfies that $p'(\lambda)\ge 0$ for all $\lambda \in [0, 1]$ and that $\int_0^1 p'(\lambda) d\lambda = 1$. Thus $p'$ is a distribution on 
$[0, 1]$ and ${\mathbb D}$ indeed maps ${\cal P}$ to ${\cal P}$.

\begin{thm}
\label{thm:MixDATequalExp}
 Let $(x, x')_1^K$ be a sequence of object pairs on which MixUp with policy $P^{\rm Mix}$ and DAT with policy $P^{\rm DAT}$ will apply independently.  If $(x, x')_1^K$ is symmetric and $P^{\rm DAT} = {\mathbb D}(P^{\rm Mix})$, then
\[
{\cal L}^{\rm Mix}\left((x, x')_1^K \right) = {\cal L}^{\rm DAT}\left((x, x')_1^K \right) 
\]
\end{thm}

Under the condition of Theorem \ref{thm:MixDATequalExp},  we see that the overall loss in MixUp and that in DAT are identical {\em in expectation}. That is, the two optimization problems would be the same if the randomness induced by their respective probabilistic policies were averaged out.  But due to Lemma \ref{lem:mixUpConverge1} and Lemma \ref{lem:DATConverge1}, the overall losses of the two schemes converge in probability to their respective expectations; they must thus be close to each other for large $K$. In fact, the following theorem is easy to prove.

\begin{thm}
\label{thm:MixDATclose} Let $(X, X')_1^\infty$ be drawn i.i.d. from $Q$. Let $\Lambda_1^\infty$ be drawn i.i.d. from $P^{\rm Mix}$ and applied to  $(X, X')_1^\infty$ using MixUp. 
Let $\Upsilon_1^\infty$ be drawn i.i.d. from $P^{\rm DAT}$ and applied to  $(X, X')_1^\infty$ using DAT. If $P^{\rm DAT} = {\mathbb D}\left(P^{\rm Mix} \right)$, 
then 
$
\left|{\cal L}^{\rm Mix}\left((X, X')_1^K, \Lambda_1^K \right) - 
{\cal L}^{\rm DAT}\left((X, X')_1^K, \Upsilon_1^K \right) \right| \stackrel{\rm p}{\longrightarrow} 0, ~{\rm as} ~ K\rightarrow \infty
$
\end{thm} 

We note that in this theorem, we no long require that the sequence $(X, X')_1^K$ has a symmetric empirical distribution. This is because for sufficiently large $K$, the empirical distribution 
becomes arbitrarily close to $Q$, which is symmetric by  definition. This will only cause a diminishing difference in ${\cal L}^{\rm Mix}\left((X, X')_1^K \right)$ and ${\cal L}^{\rm DAT}\left((X, X')_1^K \right)$.  Then by invoking Lemma \ref{lem:mixUpConverge1} and Lemma \ref{lem:DATConverge1}, it is possible to obtain a lower bound 
of ${\rm Pr}\left[
\left|{\cal L}^{\rm Mix}\left((X, X')_1^K, \Lambda_1^K \right) - 
{\cal L}^{\rm DAT}\left((X, X')_1^K, \Upsilon_1^K \right) \right|<\epsilon\right]$ that approaches $1$ as $K$ increases.  The proof is somewhat more technical, which we skip. 

This theorem suggests that at large $K$, as long as $(x, x')_1^K$ is drawn i.i.d. from a symmetric distribution on ${\cal D}\times {\cal D}$ and $P^{\rm DAT}={\mathbb D}\left(P^{\rm Mix}\right)$, the overall loss functions of the two optimization problems have very close values at each model parameter configuration $\theta$.  Thus one may argue that at large $K$, the two optimization problems have nearly the same loss landscapes and thus have nearly the same solution. We believe that it is possible to establish sharper theorems to more rigorously support such a claim, but this merits a lengthier discussion than can be apportioned in this paper.

The fact that MixUp is nearly the same as DAT under certain adversarial policies allows us to explain the effectiveness of MixUp in terms of adversarial training. In adversarial training, a data point is moved away from its original location and yet keeps its label to prevent the model from confining the class boundaries to be very close to the data points. This allows the model to generalize better to unseen regions in the data manifold, thereby preventing overfitting. 

From Theorems 
\ref{thm:MixDATequalExp} and \ref{thm:MixDATclose}, one may conclude that there exists at least a class of DAT schemes that is equivalent to MixUp in the expected loss or close to the MixUp in the original overall loss. These DAT schemes are those having an adversarial policy in the form of ${\mathbb D}\left(p\right)$, where $p$ is any distribution on $[0, 1]$.
 Then some questions naturally arise. Are there other DAT schemes that do not correspond to MixUp in this way? If there are, do they serve as more effective regularization schemes than MixUp? 

Before we answer these questions, we need to consider a generalization of MixUp. 


\subsection{Untied MixUp}

Let $\gamma$ be a function mapping $[0,1]$ to $[0, 1]$.  The scheme of Untied MixUp is exactly the same as MixUp, except that we replace $\ell^{\rm Mix}$ by another function $\ell^{\rm uMix}$ which also depends on $\gamma$ and is defined as
\begin{equation}
\ell^{\rm uMix}(x, x', \lambda, \gamma): = \ell 
\left(
F(\lambda x + \overline{\lambda}x'), \gamma(\lambda)\varphi(t(x)) + \overline{\gamma(\lambda)}\varphi(t(x'))
\right)
\end{equation}
The corresponding overall loss function ${\cal L}^{\rm Mix}\left((x, x')_1^K, \lambda_1^K\right)$ and expected overall function ${\cal L}^{\rm Mix}\left((x, x')_1^K\right)$ are denoted 
by ${\cal L}^{\rm uMix}\left((x, x')_1^K, \lambda_1^K, \gamma\right)$ and ${\cal L}^{\rm uMix}\left((x, x')_1^K, \gamma\right)$ respectively. 

\begin{lem}
\label{lem:key2}
For any $(x, x') \in {\cal D} \times {\cal D}$ and any $\lambda \in [0, 1]$,
\begin{eqnarray*}
\ell^{\rm uMix}(x, x', \lambda, \gamma)  & = & \gamma(\lambda) \ell^{\rm DAT} (x \rightarrow x', \lambda) + \overline{\gamma(\lambda)} \ell^{\rm DAT} (x' \rightarrow x,\overline{\lambda}) \\
& =  &
{\mathbb E}_{W|\lambda}  \left\{
W \cdot  \ell^{\rm DAT} (x \rightarrow x', \lambda) + (1-W) \cdot \ell^{\rm DAT} (x' \rightarrow x,\overline{\lambda})
\right\}
\end{eqnarray*}
where $W|\lambda$ is a  ${\rm Ber}(\gamma(\lambda))$) random variable.
\end{lem}

Lemma \ref{lem:key2} generalizes Lemma \ref{lem:key} to Untied MixUp. Specifically, in MixUp, the Bernoulli parameter $W|\lambda$ must be  the same as the mixing policy $\lambda$. In Untied MixUp, this parameter is ``untied'' from the mixing policy, and can be any $\gamma(\lambda)$. We will refer to $\gamma$ as the {\em weighting function}.
Then an Untied MixUp scheme is specified both by the mixing policy $P^{\rm Mix}$ and the weighting function $\gamma$.


\begin{lem}
\label{lem:uMixUpConverge1}
For any fixed infinite sequence $(x, x')_1^\infty$ and any infinite sequence of i.i.d. random variables $\Lambda_1^\infty$ drawn from $P^{\rm Mix}$, let 
${\cal L}^{\rm Mix}\left((x, x' )_1^K, \Lambda_1^K \right)$ be defined according to (\ref{eq:Lmix}), with the first $K$ elements of $(x, x')_1^\infty$ and the first $K$ elements of $\Lambda_1^\infty$ as input. Let
\[
\delta_{\rm uMix}: = \max_{(x, x') \in {\cal D} \times {\cal D}}  \sup_{(\lambda, \lambda') \in [0, 1]\times [0, 1]} \left| \ell^{\rm uMix}(x, x', \lambda, \gamma) - \ell^{\rm uMix}(x, x', \lambda', \gamma) \right| .
\]
Then  for any $\epsilon >0$, 
$
{\rm Pr}\left[
\left|
{\cal L}^{\rm Mix}\left(
(x, x')_1^K, \Lambda_1^K
\right) -  {\cal L}^{\rm Mix}\left(
\left(x, x'\right)_1^K
\right) 
\right| \ge \epsilon
\right]
<
2 \exp \left( -  \frac{2\epsilon^2}{\delta_{\rm uMix}^2} \cdot K\right).
$
\end{lem}

The proof of this lemma follows similarly to that of Lemma \ref{lem:mixUpConverge1}. 


\subsection{Relationship between Untied MixUp and DAT}

Let ${\cal F}$ denote the space of all functions mapping $[0, 1]$ to $[0, 1]$.  Each configuration in ${\cal P}\times {\cal F}$ defines an Untied MixUp scheme. 

We now define ${\mathbb U}$, which maps a DAT scheme to a Untied MixUp scheme.  Specifically ${\mathbb U}$ is a map from ${\cal P}$ to ${\cal P}\times {\cal F}$ such that for any $p\in {\cal P}$, ${\mathbb U}(p)$ is a configuration $(p', g)\in  {\cal P}\times {\cal F}$, where
\[
p'(\lambda)  := \frac{1}{2}\left( 
p(\lambda) + p(1-\lambda)
\right)~
{\rm and} ~
g(\lambda) :=  \frac{p(\lambda)}{p(\lambda) + p(1-\lambda)}
\]

\begin{thm}
\label{thm:uMixDATequalExp}
 Let $(x, x')_1^K$ be a sequence of object pairs on which an Untied MixUp scheme specified by $(P^{\rm Mix}, \gamma)$ and a DAT scheme with policy $P^{\rm DAT}$ will apply independently.  If $(x, x')_1^K$ is symmetric and $\left(P^{\rm Mix}, \gamma\right) = {\mathbb U}(P^{\rm DAT})$, then
$
{\cal L}^{\rm uMix}\left((x, x')_1^K, \gamma\right) = {\cal L}^{\rm DAT}\left((x, x')_1^K \right). 
$
\end{thm}

Note that in the definition of ${\mathbb U}$,  $g(\lambda)$ is undefined at the values of $\lambda$ for which the denominator is zero. This may result in $\gamma(\lambda)$ in the theorem being undefined for some $\lambda \in [0, 1]$. But even in this case, the theorem still holds true. This is because, those $\lambda$ for which $\gamma(\lambda)$ is undefined never gets drawn in the DAT scheme, thus creating no problem.

We now define another map ${\mathbb D}_{\rm u}$ that maps an Untied MixUp scheme to a DAT scheme.  Specifically ${\mathbb D}_u$ is a map from  ${\cal P}\times {\cal F}$ 
to ${\cal P}$  such that for any $(p, g)\in {\cal P}\times {\cal F}$, ${\mathbb D}_{\rm u}(p, g)$ is a configuration $p'\in  {\cal P}$, where
\[
p'(\lambda) := \left(
g(\lambda) p(\lambda)  
+ \overline{g(\overline{\lambda})} p(1-\lambda) 
\right)
\]
It is easy to verify that $\int_0^1 p'(\lambda) d\lambda = 1$. Thus $p'$ is indeed a distribution in ${\cal P}$ and ${\mathbb D}_{\rm u}$ is well defined.

\begin{thm}
\label{thm:uMixDATequalExp2}
 Let $(x, x')_1^K$ be a sequence of object pairs on which an Untied MixUp scheme specified by $(P^{\rm Mix}, \gamma)$ and a DAT scheme with policy $P^{\rm DAT}$ will apply independently.  If $(x, x')_1^K$ is symmetric and $P^{\rm DAT}={\mathbb D}_{\rm u}\left(P^{\rm Mix}, \gamma\right)$, then
$
{\cal L}^{\rm uMix}\left((x, x')_1^K, \gamma\right) = {\cal L}^{\rm DAT}\left((x, x')_1^K \right). 
$
\end{thm}

When the object-pair data is symmetric, Theorems \ref{thm:uMixDATequalExp} and \ref{thm:uMixDATequalExp2} suggest that for every Untied MixUp scheme, there is a DAT scheme giving rise to the same expected overall loss and that  the converse also holds.  Thus the family of Untied MixUp schemes and the family of DAT schemes are ``equivalent'' in their expected overall losses. This equivalence also implies the following result. 

\begin{thm}
\label{thm:uMixDATclose} Let $(X, X')_1^\infty$ be drawn i.i.d. from $Q$. On this object-pair data, an Untied MixUp scheme specified by $(P^{\rm Mix}, \gamma)$ and a DAT scheme specified by $P^{\rm DAT}$ will apply.  In the Untied MixUp scheme, let  $\Lambda_1^\infty$ be drawn i.i.d. from $P^{\rm Mix}$; in the DAT scheme, let $\Upsilon_1^\infty$ be drawn i.i.d. from $P^{\rm DAT}$. If 
$P^{\rm DAT}={\mathbb D}_{\rm u}\left(P^{\rm Mix}, \gamma\right)$ or $\left(P^{\rm Mix}, \gamma\right) = {\mathbb U}(P^{\rm DAT})$, then 
\[
\left|{\cal L}^{\rm Mix}\left((X, X')_1^K, \Lambda_1^K , \gamma\right) - 
{\cal L}^{\rm DAT}\left((X, X')_1^K, \Upsilon_1^K \right) \right| \stackrel{\rm p}{\longrightarrow} 0, ~{\rm as} ~ K\rightarrow \infty
\]
\end{thm} 

The equivalence between the two families of schemes also indicates that there are indeed DAT schemes that do not correspond to a MixUp scheme, answering a question above. These DAT schemes correspond to Untied MixUp scheme beyond the standard MixUp. The relationship between MixUp, DAT and Untied MixUp is shown in Figure \ref{fig:diagram}.


\section{Experiments}

\subsection{Experiment Setup and Implementation}

We consider an image classification task on the Cifar10 and Cifar100 data set. 
The baseline classifier chosen is PreActResNet18 (see \cite{github}), noting the same choice is made by the authors of Mixup\cite{zhang2017mixup}.

Both MixUp and Untied MixUp are considered in the experiments. The MixUp policies are chosen as Beta distribution $B(\alpha, \beta)$.
The Untied MixUp policy is taken as  ${\mathbb U}(B(\alpha, \beta))$.




Two target-linear loss functions are essayed: cross-entropy (CE) loss and the negative-cosine (CE) loss as defined earlier. We implement CE loss similarly to previous works, which use CE loss to implement the baseline model. In our implementation of the NC loss model, for each label $y$, $\varphi(y)$ is mapped to a randomly selected unit-length vector of dimension $d$ and fixed during training; the feature map of the original PreActResNet18 is linearly transformed to a $d$-dimensional vector. The dimension $d$ is chosen as 300 for Cifar10 and 700 for Cifar100.

Our implementation of MixUp and Untied MixUp improves upon the published implementation from the original authors of MixUp \cite{zhang2017mixup}.  
For example, the original authors' implementation samples only one $\lambda$ per mini-batch, giving rise to unnecessarily higher stochasticity of the gradient signal. Our implementation samples $\lambda$ independently for each sample. Additionally, the original code combines inputs by mixing a mini-batch of samples with a shuffled version of itself. This approach introduces a dependency between sampled pairs and again increases the stochasticity of training. 
Our implementation creates two shuffled copies
of the entire training dataset prior to each epoch, pairs them up, and then splits them into mini-batches. This gives a closer approximation to i.i.d. sampling and makes training smoother.
While these implementation improvements have merit on their own, they do not provide a theoretical leap in understanding, and so we do not quantify their impact in our results analysis.

All models examined are trained using mini-batched backpropagation, for 200 epochs.

\subsection{Results}

We sweep over the policy space of MixUp and Untied MixUp. For MixUp, it is sufficient to consider distribution $P^{\rm Mix}$ to be symmetric about $0.5$. Thus we consider only consider $P^{\rm Mix}$  in the form of ${\rm B}(\alpha, \alpha)$, and scan through a single parameter $\alpha$ systematically. Since the policy of Untied MixUp is in the form of ${\mathbb U}({\rm B}(\alpha, \beta))$, searching through $(\alpha, \beta)$ becomes more difficult. Thus our policy search for Untied MixUp is restricted to an {\em ad hoc} heuristic search.  For this reason, the found best policy for Untied MixUp might be quite far from the true optimal. 


The main results of our experiments are given in Tables 1 and 2.  As shown in the tables, each setting is run multiple times. In fact, we began our experimentation using 50 runs for each policy setting, and progressively reduced to 6 runs as we hit computation resource limitations. This explains the varying number of runs in the tables.

For each run,  we compute the error rate in a run as the average test error rate over the final 10 epochs. The estimated mean (``MEAN'')) performance of a setting is computed as the average of the error rates over all runs for the same setting. The 95\%-confidence interval (``ConfInt'') for the estimated mean performance is also computed and shown in the table.


On Cifar100, one can conclude with confidence that the found Untied MixUp presents a better average performance that the best found MixUp. That suggests that generalizing MixUp to the untied offers additional room for more effective regularization.  On Cifar10, the empirical improvement brought by Untied MixUp is smaller. This may be because Cifar10 is a simpler dataset for which the highly optimized baseline and MixUp models leave less room for improvement. 

The results show empirically that MixUp and Untied MixUp both work on the NC loss models. This validates our generalization of MixUp (and Untied MixUp) to models built with target linear losses.  The NC loss models appear to perform worse that the CE loss models. We are unsure whether this is because the baseline model isn't optimized for the NC loss model, or due to some inherent limitation in training with the NC loss.

\begin{table}[ht!]
    \centering
    \begin{tabular}{|c|c|c|c|c|}
    \hline 
    model & policy & runs & MEAN & ConfInt\\
    \hline
    baseline-CE & $-$ & 50 & 5.533\%, & 0.033\% \\
    \hline
    mixUp-CE & ${\rm B(0.9,0.9)}$ & 100 & 4.172\% & 0.023\% \\
    \hline
    uMixUp-CE & ${\mathbb U}({\rm B(2.2,0.9)})$ & 100 & 4.175\% & 0.0247\% \\
    \hline
    \hline
    baseline-NC & $-$ & 12 & 5.624\% & 0.078\% \\
    \hline
    mixUp-NC & ${\rm B(0.9,0.9)}$ & 12 &  4.584\% & 0.062\% \\
    \hline
    uMixUp-NC & ${\mathbb U}({\rm B(2.2,0.9)})$ & 12 & 4.544\% & 0.082\% \\
    \hline
    \end{tabular}
    \caption{Test error rate on CIFAR10. }
\label{tab:cifar10_mixUpVsUntiedMixUp}
\end{table}

\begin{table}[ht!]
    \centering
    \begin{tabular}{|c|c|c|c|c|}
    \hline 
    model & policy & runs & MEAN & ConfInt\\
    \hline
    baseline-CE & $-$ & 50 & 24.932\% & 0.075\% \\
    \hline
    mixUp-CE & ${\rm B(0.9,0.9)}$ & 100 & 21.942\% & 0.040\% \\
    \hline
    uMixUp-CE & ${\mathbb U}({\rm B(1.4,0.7)})$ & 100 & 21.863\% & 0.057\%  \\
    \hline
    \hline
    baseline-NC & $-$ & 12  & 25.501\% & 0.237\% \\
    \hline
    mixUp-NC & ${\rm B(0.9,0.9)}$ & 12 & 24.418\% & 0.175\% \\
    \hline
    uMixUp-NC & ${\mathbb U}{\rm (B(1.4,0.7)} )$& 12  & 23.926\% & 0.151\%  \\
    \hline
    \end{tabular}
    \caption{Test error rate on CIFAR100. }
    \label{tab:cifar10_mixUpVsUntiedMixUp}
\end{table}

\section{Concluding Remarks}

This paper establishes a connection between MixUp and adversarial training.  This connection allows for a better understanding of the working mechanism of MixUp as well as a generalization of MixUp to a wider family, namely Untied MixUp. Despite the development in this work, it is the authors' belief that the current designs of MixUp and Untied MixUp are far from optimal. In particular, we believe a better design should allow individualized policy for each training pair. How this can be done remains open at this time.

\bibliographystyle{alpha}
\bibliography{ref}

\begin{thebibliography}{10}

\bibitem{CloserLookAtMemorization_Bengio}
Devansh Arpit, Stanislaw~K. Jastrzebski, Nicolas Ballas, David Krueger,
  Emmanuel Bengio, Maxinder~S. Kanwal, Tegan Maharaj, Asja Fischer, Aaron~C.
  Courville, Yoshua Bengio, and Simon Lacoste{-}Julien.
\newblock A closer look at memorization in deep networks.
\newblock In {\em Proceedings of the 34th International Conference on Machine
  Learning, {ICML} 2017, Sydney, NSW, Australia, 6-11 August 2017}, pages
  233--242, 2017.

\bibitem{athalye2018obfuscated}
Anish Athalye, Nicholas Carlini, and David Wagner.
\newblock Obfuscated gradients give a false sense of security: Circumventing
  defenses to adversarial examples.
\newblock {\em arXiv preprint arXiv:1802.00420}, 2018.

\bibitem{Bert}
Jacob Devlin, Ming{-}Wei Chang, Kenton Lee, and Kristina Toutanova.
\newblock {BERT:} pre-training of deep bidirectional transformers for language
  understanding.
\newblock {\em CoRR}, abs/1810.04805, 2018.

\bibitem{goodFellowAdv}
Ian~J Goodfellow, Jonathon Shlens, and Christian Szegedy.
\newblock Explaining and harnessing adversarial examples.
\newblock {\em arXiv preprint arXiv:1412.6572}, 2014.

\bibitem{adaMixUp}
Hongyu Guo, Yongyi Mao, and Richong Zhang.
\newblock Mixup as locally linear out-of-manifold regularization.
\newblock {\em arXiv preprint arXiv:1809.02499}, 2018.

\bibitem{resNet}
Kaiming He, Xiangyu Zhang, Shaoqing Ren, and Jian Sun.
\newblock Deep residual learning for image recognition.
\newblock In {\em Proceedings of the IEEE conference on computer vision and
  pattern recognition}, pages 770--778, 2016.

\bibitem{he2018decision}
Warren He, Bo~Li, and Dawn Song.
\newblock Decision boundary analysis of adversarial examples.
\newblock In {\em 6th International Conference on Learning Representations,
  {ICLR} 2018, Vancouver, BC, Canada, April 30 - May 3, 2018, Conference Track
  Proceedings}, 2018.

\bibitem{WeightDecay}
Anders Krogh and John~A. Hertz.
\newblock A simple weight decay can improve generalization.
\newblock In {\em Advances in Neural Information Processing Systems 4, {NIPS}
  Conference, Denver, Colorado, USA, December 2-5, 1991}, pages 950--957, 1991.

\bibitem{github}
Kuang Liu.
\newblock URL \url{https://github.com/kuangliu/pytorch-cifar}, 2017.

\bibitem{mcdiarmid1989method}
Colin McDiarmid.
\newblock On the method of bounded differences.
\newblock {\em Surveys in combinatorics}, 141(1):148--188, 1989.

\bibitem{miyato2016adversarial}
Takeru Miyato, Andrew~M Dai, and Ian Goodfellow.
\newblock Adversarial training methods for semi-supervised text classification.
\newblock {\em arXiv preprint arXiv:1605.07725}, 2016.

\bibitem{shaham2018understanding}
Uri Shaham, Yutaro Yamada, and Sahand Negahban.
\newblock Understanding adversarial training: Increasing local stability of
  supervised models through robust optimization.
\newblock {\em Neurocomputing}, 307:195--204, 2018.

\bibitem{dropout}
Nitish Srivastava, Geoffrey~E. Hinton, Alex Krizhevsky, Ilya Sutskever, and
  Ruslan Salakhutdinov.
\newblock Dropout: a simple way to prevent neural networks from overfitting.
\newblock {\em Journal of Machine Learning Research}, 15(1):1929--1958, 2014.

\bibitem{szegedy2013intriguing}
Christian Szegedy, Wojciech Zaremba, Ilya Sutskever, Joan Bruna, Dumitru Erhan,
  Ian Goodfellow, and Rob Fergus.
\newblock Intriguing properties of neural networks.
\newblock {\em arXiv preprint arXiv:1312.6199}, 2013.

\bibitem{manifoldMixUp}
Vikas Verma, Alex Lamb, Christopher Beckham, Amir Najafi, Aaron Courville,
  Ioannis Mitliagkas, and Yoshua Bengio.
\newblock Manifold mixup: Learning better representations by interpolating
  hidden states.
\newblock 2018.

\bibitem{wideResNet}
Sergey Zagoruyko and Nikos Komodakis.
\newblock Wide residual networks.
\newblock In {\em Proceedings of the British Machine Vision Conference 2016,
  {BMVC} 2016, York, UK, September 19-22, 2016}, 2016.

\bibitem{RethinkingGeneralization}
Chiyuan Zhang, Samy Bengio, Moritz Hardt, Benjamin Recht, and Oriol Vinyals.
\newblock Understanding deep learning requires rethinking generalization.
\newblock In {\em 5th International Conference on Learning Representations,
  {ICLR} 2017, Toulon, France, April 24-26, 2017, Conference Track
  Proceedings}, 2017.

\bibitem{zhang2017mixup}
Hongyi Zhang, Moustapha Cisse, Yann~N Dauphin, and David Lopez-Paz.
\newblock mixup: Beyond empirical risk minimization.
\newblock {\em arXiv preprint arXiv:1710.09412}, 2017.
\newblock URL \url{https://github.com/facebookresearch/mixup-cifar10}.

\end{thebibliography}

\clearpage

\section*{Supplementary Materials}

%

\subsection*{Proof of Lemma 2:}

For any given $\lambda_1^K\in [0, 1]^K$ and any of its modified version $u_1^K\in [0, 1]^K$ which differs from $\lambda_1^K$ in exactly one location, it can be verify, 
following the definition of $\delta_{\rm Mix}$, that
\[
\left|
{\cal L}^{\rm Mix}\left(
(x, x')_1^K, \lambda_1^K
\right) - 
{\cal L}^{\rm Mix}\left(
(x, x')_1^K, u_1^K
\right) 
\right| \le \delta_{\rm Mix}/K
\]
Since $\Lambda_1, \Lambda_2, \ldots \Lambda_K$ are independent and by McDiarmid Inequality \cite{mcdiarmid1989method}, it follows that  for any $\epsilon >0$, 
\[
{\rm Pr}\left[
\left|
{\cal L}^{\rm Mix}\left(
(x, x')_1^K, \Lambda_1^K
\right) -  {\cal L}^{\rm Mix}\left(
\left(x, x'\right)_1^K
\right) 
\right| \ge \epsilon
\right]
<
2 \exp \left( - \frac{2\epsilon^2}{K\cdot \left( \delta_{\rm Mix}/K\right)^2}\right) 
\]
which proves the lemma \hfill $\Box$




\subsection*{Proof of Theorem 1:}

\begin{eqnarray*}
{\cal L}^{\rm Mix} \left((x, x')_1^K \right) 
\!\!\!\!\!\!\!\!\!\!\!\!\!& = & 
\frac{1}{K}\sum\limits_{k=1}^K  {\mathbb E}_{\lambda_k \sim P^{\rm Mix}} \ell^{\rm Mix}(x_k,  x'_k, \lambda_k)\\
&\stackrel{({\rm Lemma}~ \ref{lem:key})}{=} &
\frac{1}{K}
\left\{
\sum\limits_{k=1}^K \int \lambda P^{\rm Mix}(\lambda) \ell^{\rm DAT}(x_k\rightarrow x'_k, \lambda) d\lambda
+ 
\sum\limits_{k=1}^K \int \overline{\lambda} P^{\rm Mix}(\lambda) \ell^{\rm DAT}(x'_k\rightarrow x_k, \overline{\lambda}) d\lambda
\right\}\\
& \stackrel{({\rm symmetry})}= & 
\frac{1}{K}
\left\{
\sum\limits_{k=1}^K \int \lambda P^{\rm Mix}(\lambda) \ell^{\rm DAT}(x_k\rightarrow x'_k, \lambda) d\lambda
+ 
\sum\limits_{k=1}^K \int \overline{\lambda} P^{\rm Mix}(\lambda) \ell^{\rm DAT}(x_k\rightarrow x'_k, \overline{\lambda}) d\lambda
\right\}\\
& = & 
\frac{1}{K}
\left\{
\sum\limits_{k=1}^K \int \lambda P^{\rm Mix}(\lambda) \ell^{\rm DAT}(x_k\rightarrow x'_k, \lambda) d\lambda
+ 
\sum\limits_{k=1}^K \int \lambda P^{\rm Mix}(1-\lambda) \ell^{\rm DAT}(x_k\rightarrow x'_k, \lambda) d\lambda
\right\}\\
& = & 
\frac{1}{K}
\sum\limits_{k=1}^K 
\int \lambda \left(
P^{\rm Mix}(\lambda) +P^{\rm Mix}(1-\lambda)
\right)
\ell^{\rm DAT}(x_k\rightarrow x'_k, \lambda) d\lambda
\\
& = & 
\frac{1}{K}
\sum\limits_{k=1}^K 
\int P^{\rm DAT}(\lambda)
\ell^{\rm DAT}(x_k\rightarrow x'_k, \lambda) d\lambda
\\
& = & 
\frac{1}{K}
\sum\limits_{k=1}^K 
{\mathbb E}_{\lambda_k\sim P^{\rm DAT}} \ell^{\rm DAT}(x_k\rightarrow x'_k, \lambda_k)\\
& = & {\cal L}^{\rm DAT}\left( (x, x')_1^K\right)
\end{eqnarray*}
This completes the proof. \hfill $\Box$





\subsection* {Proof of Theorem 3:} 
\begin{eqnarray*}
{\cal L}^{\rm uMix}
\left(
(x, x')_1^K, \gamma
\right)
& \!\!\!\!\!\! := & \!\!\!\!\!\!
 \frac{1}{K}
 \sum\limits_{k=1}^K 
 {\mathbb E}_{\lambda \sim P^{\rm Mix}} 
 \left\{
 \gamma(\lambda) \ell^{\rm DAT}(x_k \rightarrow x'_k, \lambda)
 + \overline{\gamma(\lambda)} \ell^{\rm DAT}(x'_k \rightarrow x_k, \overline{\lambda})
 \right\} \\
& = & \!\!\!\!\!\!
 \frac{1}{K}
 \sum\limits_{k=1}^K 
 \int
 \left(
 \gamma(\lambda) P^{\rm Mix}(\lambda)
  \ell^{\rm DAT}(x_k \rightarrow x'_k, \lambda) \right. 
\left.
 + 
\overline{ \gamma(\lambda)} P^{\rm Mix}(\lambda)
 \ell^{\rm DAT}(x'_k \rightarrow x_k,  \overline{\lambda})
 \right)
 d\lambda \\
& = & \!\!\!\!\!\!
 \frac{1}{K}
 \sum\limits_{k=1}^K 
 \int
 \left(
 \frac{1}{2}P^{\rm DAT}(\lambda)
  \ell^{\rm DAT}(x_k \rightarrow x'_k, \lambda) \right. 
\left.
 + 
\frac{1}{2}P^{\rm DAT}(\overline{\lambda})
 \ell^{\rm DAT}(x'_k \rightarrow x_k,  \overline{\lambda})
 \right)
 d\lambda \\
& = & \!\!\!\!\!\!
 \frac{1}{2K}
 \left(
 \sum\limits_{k=1}^K 
 \int
 P^{\rm DAT}(\lambda)
  \ell^{\rm DAT}(x_k \rightarrow x'_k, \lambda) d\lambda
 + 
 \sum\limits_{k=1}^K 
 \int
 P^{\rm DAT}(\overline{\lambda})
 \ell^{\rm DAT}(x'_k \rightarrow x_k,  \overline{\lambda})
 d\lambda
 \right) \\
& \stackrel{(a)}{=} & \!\!\!\!\!\!
 \frac{1}{2K}
 \left(
 \sum\limits_{k=1}^K 
 \int
 P^{\rm DAT}(\lambda)
  \ell^{\rm DAT}(x_k \rightarrow x'_k, \lambda) d\lambda
 + 
 \sum\limits_{k=1}^K 
 \int
 P^{\rm DAT}(\lambda)
 \ell^{\rm DAT}(x'_k \rightarrow x_k,  \lambda)
 d\lambda
 \right) \\
& \stackrel{(b)}{=} & \!\!\!\!\!\!
 \frac{1}{2K}
 \left(
 \sum\limits_{k=1}^K 
 \int
 P^{\rm DAT}(\lambda)
  \ell^{\rm DAT}(x_k \rightarrow x'_k, \lambda) d\lambda
 + 
 \sum\limits_{k=1}^K 
 \int
 P^{\rm DAT}(\lambda)
 \ell^{\rm DAT}(x_k \rightarrow x'_k,  \lambda)
 d\lambda
 \right) \\
& = & \!\!\!\!\!\!
 \frac{1}{K}
 \sum\limits_{k=1}^K 
 \int
 P^{\rm DAT}(\lambda)
  \ell^{\rm DAT}(x_k \rightarrow x'_k, \lambda) d\lambda\\
 & = & \!\!\!\!\!\!
{\cal L}^{\rm DAT}
\left(
(x, x')_1^K
\right)
 \end{eqnarray*}
where (a) is due to a change of variable in the integration,  (b) is due to the symmetry of $(x, x')_1^K$.  \hfill $\Box$

\subsection*{ Proof of Theorem 4}
\begin{eqnarray*}
{\cal L}^{\rm uMix}_K\left((x, x')_1^K, \gamma \right) 
& = & 
\frac{1}{K} {\mathbb E}_{\lambda \sim P^{\rm Mix}} 
\sum\limits_{k=1}^K \left(
\gamma(\lambda) \ell^{\rm DAT}(x_k\rightarrow x'_k, \lambda)
+ \overline{\gamma(\lambda)} \ell^{\rm DAT}(x'_k\rightarrow x_k, \overline{\lambda})
\right) \\
& = & 
\frac{1}{K} 
\left(
{\mathbb E}_{\lambda \sim P^{\rm Mix}} 
\sum\limits_{k=1}^K 
\gamma(\lambda) \ell^{\rm DAT}(x_k\rightarrow x'_k, \lambda)
+ {\mathbb E}_{\lambda \sim P^{\rm Mix}} 
\sum\limits_{k=1}^K \overline{\gamma(\lambda)} \ell^{\rm DAT}(x'_k\rightarrow x_k, \overline{\lambda})
\right) \\
&\stackrel{(a)} {=} & 
\frac{1}{K} 
\left(
{\mathbb E}_{\lambda \sim P^{\rm Mix}} 
\sum\limits_{k=1}^K 
\gamma(\lambda) \ell^{\rm DAT}(x_k\rightarrow x'_k, \lambda)
+ {\mathbb E}_{\lambda \sim P^{\rm Mix}} 
\sum\limits_{k=1}^K \overline{\gamma(\lambda)} \ell^{\rm DAT}(x_k\rightarrow x'_k, \overline{\lambda})
\right) \\
&\stackrel{(b)} {=} & 
\frac{1}{K} 
\left(
{\mathbb E}_{\lambda \sim P^{\rm Mix}} 
\sum\limits_{k=1}^K 
\gamma(\lambda) \ell^{\rm DAT}(x_k\rightarrow x'_k, \lambda)
+ {\mathbb E}_{\overline{\lambda} \sim P^{\rm Mix}} 
\sum\limits_{k=1}^K \overline{\gamma(\overline{\lambda})} \ell^{\rm DAT}(x_k\rightarrow x'_k, \lambda)
\right) \\
& = &
\frac{1}{K} \sum\limits_{k=1}^K 
\int 
\left(
\gamma(\lambda) P^{\rm Mix}(\lambda)  \ell^{\rm DAT}(x_k\rightarrow x'_k, \lambda)
+ \overline{\gamma(\overline{\lambda})} P^{\rm Mix}(1-\lambda)  \ell^{\rm DAT}(x_k\rightarrow x'_k, \lambda)
\right)
d\lambda \\
& = &
\frac{1}{K} \sum\limits_{k=1}^K 
\int  \ell^{\rm DAT}(x_k\rightarrow x'_k, \lambda)
\underbrace{\left(
\gamma(\lambda) P^{\rm Mix}(\lambda)  
+ \overline{\gamma(\overline{\lambda})} P^{\rm Mix}(1-\lambda) 
\right)}_{{\mathbb D}_{\rm u}\left(P^{\rm Mix}, \gamma\right)}
d\lambda\\
& = & \frac{1}{K} \sum\limits_{k=1}^K {\mathbb E}_{\lambda \sim P^{\rm DAT}}  \ell^{\rm DAT}(x_k\rightarrow x'_k, \lambda)\\
& = & {\cal L}^{\rm DAT} \left((x, x')_1^K\right).
\end{eqnarray*}
where (a) is due to the symmetry of $(x, x'_1)^K$, and  (b) is by a change of variable in the second term (renaming $1-\lambda$ as $\lambda$).  \hfill $\Box$

\end{document}